\documentclass[runningheads]{llncs}

\usepackage{eccv}

\usepackage{eccvabbrv}

\usepackage{graphicx}
\usepackage{booktabs}

\usepackage[accsupp]{axessibility}  % Improves PDF readability for those with disabilities.

\usepackage{caption}   % for \captionof
\usepackage{booktabs}  % 你表格在用 \toprule 等
\usepackage{graphicx}

\usepackage{hyperref}

\usepackage{orcidlink}

\usepackage{multirow}
\usepackage{graphicx}
\usepackage{booktabs}
\usepackage{multirow}
\usepackage{arydshln}

\begin{document}

% ---------------------------------------------------------------
% TODO REVIEW: Replace with your title
\title{Learning 1-Bit LiDAR-based Localization with Auxiliary Objective}
%Implicit Map

% TODO REVIEW: If the paper title is too long for the running head, you can set
% an abbreviated paper title here. If not, comment out.
\titlerunning{BiLoc}

% TODO FINAL: Replace with your author list.
% Include the authors' OCRID for the camera-ready version, if at all possible.
\author{Kaijie Yin\inst{1,2,3}\orcidlink{0009-0009-1818-7866} \and
Zhiyuan Zhang\inst{2}\orcidlink{0000-0003-3945-5638} \and
Tian Gao\inst{1}\orcidlink{0000-0002-1646-5430} \and
Wentao Zhu\inst{3}\orcidlink{0000-0002-5483-0259}  \and \\
Cheng-zhong Xu\inst{1}\orcidlink{0000-0001-9480-0356} \and
Hui Kong\inst{1}\textsuperscript{*}
}

% TODO FINAL: Replace with an abbreviated list of authors.
\authorrunning{Yin et al.}
% First names are abbreviated in the running head.
% If there are more than two authors, 'et al.' is used.

% TODO FINAL: Replace with your institution list.
\institute{University of Macau \and
Singapore Management University \and
Eastern Institute of Technology, Ningbo\
% \email{{mc35211, }@um.edu.mo}
}

\maketitle

\renewcommand{\thefootnote}{\fnsymbol{footnote}}
\footnotetext[1]{Corresponding author.}
\renewcommand{\thefootnote}{\arabic{footnote}}

\begin{abstract}

6-DoF LiDAR-based localization is a fundamental capability for autonomous systems operating in large-scale outdoor environments. Many deep-learning-based localization methods have achieved promising performance so far. However, as one of the always-on modules competing for limited on-board computational resources, the localization module is expected to consume only a small portion of the overall compute budget. Most existing learning-based methods are still too heavy for this purpose. 
In contrast, binary neural networks (BNNs) offer an appealing solution, but the 1-bit compression causes severe information loss and performance drop. In this paper, we address this challenge by proposing Binarized LiDAR-based Localization (BiLoc), the first binary neural network framework for 6-DoF LiDAR localization.
Specifically, we reinterpret the training of BNNs from the perspective of the information-bottleneck principle, aiming at retaining minimal yet sufficient representations for pose estimation while suppressing redundant variations. And we introduce an auxiliary objective that adaptively regulates information retention in the binary encoder, effectively mitigating the information loss caused by binarization. This auxiliary objective provides additional optimization signals that compensate for the limited representational capacity and the gradient mismatch inherent in BNNs.
Extensive experiments on large-scale outdoor LiDAR datasets demonstrate that BiLoc establishes a new state of the art for LiDAR localization with BNNs.
\end{abstract}
\section{Introduction}
\label{sec:intro}
% 介绍一下为什么要用localization的工作（waymo那套）。为什么要做1-bit的localization。 应用范式是在模块化自动驾驶范式中，需要定位模块尽可能轻量级。本着这个目标，我们才要做一个精简/lightweight/efficient的定位module，因此才引入1-bit。
%（传统的localization问题，显式地图制作很麻烦，需要经常定期维护。隐式地图建图相比十分方便more convient。显示地图定位过程消耗计算资源且耗时较长，如果过滤掉较多点以提升定位速度，其精度又会有显著下降。）
%调研目前有哪些轻量级的定位方案。目前基于1bit的隐式定位方案优势明显。
% 重点介绍1bit localization的挑战/难点，这才能引出我的解决方案。
%介绍我们的方法相比其他方法有什么不同。
% 后面再具体介绍我们的方法是怎么做的。

Modern autonomous driving systems~\cite{sun2020scalability, victor2023safety} are typically designed in a modular paradigm where perception, localization, planning, and control are implemented as loosely coupled components. Among these modules, localization plays a safety-critical and always-on role, as accurate pose estimation is required continuously for downstream decision making. Unlike perception backbones that can be selectively activated or offloaded, the localization module must operate under strict latency, power, and memory constraints. These constraints make lightweight and efficient localization a fundamental requirement in autonomous systems.

Traditional LiDAR localization methods~\cite{wolcott2015fast, Yin2023ASO} rely on explicit map construction and point-to-map registration, incurring high computational cost and frequent map maintenance. With the rise of deep neural networks, learning-based approaches have achieved significant progress. These methods can be broadly categorized into Scene Coordinate Regression (SCR)~\cite{brachmann2017dsac, brachmann2023accelerated, yang2025raloc} and Absolute Pose Regression (APR)~\cite{wang2021pointloc, wang2023hypliloc, li2024diffloc}. SCR predicts point-to-scene correspondences followed by a 6-DoF pose estimation via classical registration~\cite{Fischler1981Ransac}. In contrast, APR directly regresses the global pose in an end-to-end manner without explicit map matching.
However, existing APR models remain computationally expensive for deployment on resource-constrained edge devices, motivating more aggressive model compression strategies.
\begin{figure}[t]
\centering
\includegraphics[width=4.7in, keepaspectratio]{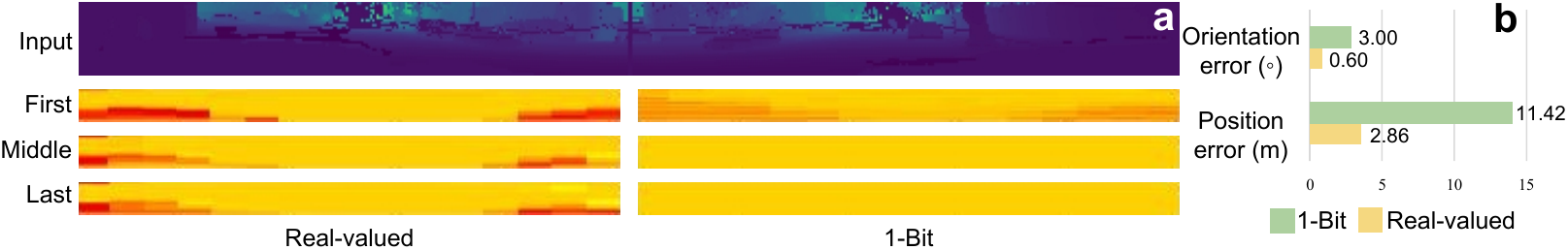}
\caption{
$\text{(a)}$ Visualization of the information loss of real-valued ViT-S/16 and ReActNet-based ViT-S/16 on the Oxford Radar RobotCar dataset~\cite{barnes2020oxford} at the first, middle, and last blocks~\cite{cheng2020explaining}. Darker colors indicate less discarded information, showing higher importance for LiDAR localization. $\text{(b)}$ Comparison of localization accuracy of full-precision ViT-S/16 and ReActNet-based ViT-S/16 on the 18-14-14-42 test split of the Oxford Radar RobotCar dataset.}
\label{fig: teaser}
\end{figure}

Existing compression techniques mainly include pruning~\cite{grainge2024structured, he2017channel}, knowledge distillation~\cite{shi2022improved, hinton2015distilling}, quantization~\cite{grainge2024design, liu2023oscillation, jacob2018quantization}, and compact network design~\cite{luo2025bevplace++, sandler2018mobilenetv2}. Model quantization is a promising solution, which significantly reduces memory usage and computational cost by quantizing weights and/or activations into low-bit integers. 
The most extreme form of quantization is binarization~\cite{bnn, birealnet, qin2020forward}, which encodes both weights and activations using 1-bit representations. BNNs replace arithmetic operations with efficient logic operations (XNOR and PopCount), achieving up to $\text{32×}$ memory compression and $\text{58×}$ computational savings~\cite{rastegari2016xnor}. These advantages make binarization attractive for localization modules that face limited-resource challenges.
However, directly applying existing binarization techniques to LiDAR-based localization leads to severe performance degradation. The primary limitation lies in the insufficient expressiveness of binary feature representations, which is particularly problematic for localization as a regression task requiring fine-grained information to resolve subtle pose differences.
In practice, this limitation is further aggravated by optimization difficulties during BNN training. Aggressive binarization restricts the representation space, while the non-differentiable sign function causes gradient mismatch. As a result, standard training objectives struggle to learn task-sufficient representations, leading to unstable optimization and poor localization accuracy.

\begin{figure}
\centering
\includegraphics[width=4.7in, keepaspectratio]{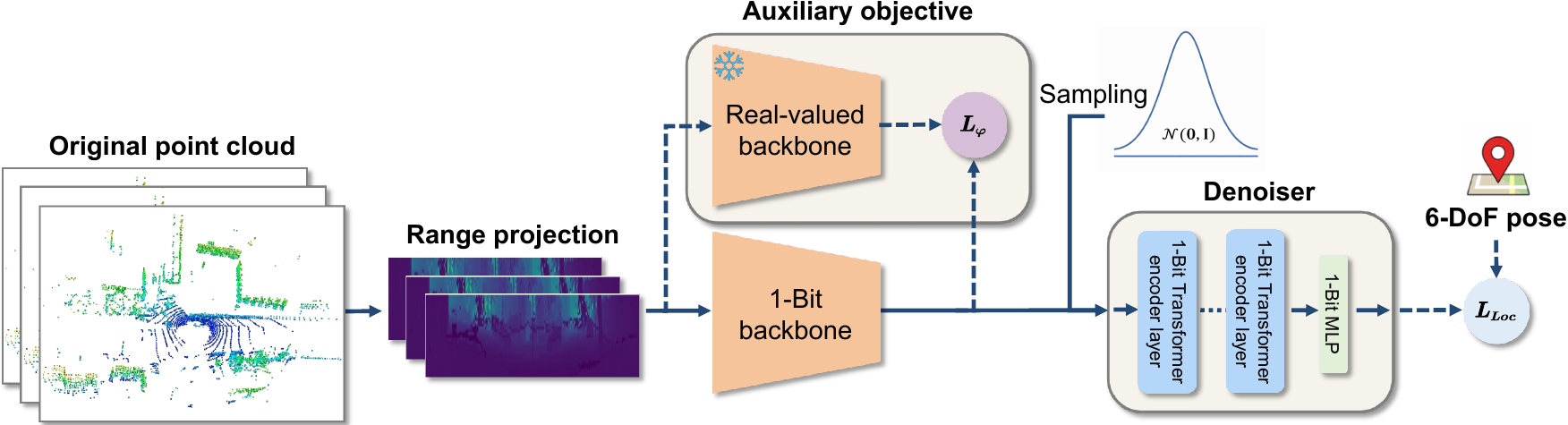}
\caption{
BiLoc framework:
The raw LiDAR point cloud is first projected into a range image that preserves geometric and depth information. The range image is fed into a binarized backbone to extract global features. These features are then passed to a fully binarized PoseDiffusion~\cite{wang2023posediffusion} decoder to predict the global pose.
During training, the backbone features are also supervised by an auxiliary objective, which provides additional optimization guidance without introducing inference overhead.
}
\label{fig: overview}
\end{figure}

To investigate this issue, we analyze BNNs for LiDAR localization from an information-theoretic perspective. We adopt an information-bottleneck based analysis method~\cite{tishby2015deep}, Quantifying Knowledge Points~\cite{zhang2022quantifying}, to compare a real-valued ViT-S/16 model~\cite{dosovitskiy2021image} with its binarized counterpart, a ReActNet-based ViT-S/16 model~\cite{reactnet, Le_2023_CVPR}. As shown in Fig.~\ref{fig: teaser}, the binarized model exhibits progressive information loss across layers during forward propagation. Compared with the real-valued network, the 1-bit model shows weaker ability to capture task-relevant information even in shallow blocks. This issue becomes more severe in deeper layers, where limited representation capacity and gradient mismatch further amplify information loss. These observations suggest that the main challenge of 1-bit localization lies not only in reduced capacity, but also in the lack of mechanisms to regulate information retention under extreme compression.

Based on these observations, we propose Binarized LiDAR-based Localization (BiLoc). Instead of training BNNs with a single objective, BiLoc formulates the learning process under the information-bottleneck principle~\cite{tishby2000information, shwartz2017opening}, aiming to retain minimal yet task-sufficient representations for pose estimation while suppressing irrelevant variations. To facilitate this objective, we introduce an auxiliary objective that performs feature distillation using representations from an offline real-valued model as a reference distribution. This additional supervision compensates for the limited representational capacity and gradient mismatch of BNNs, guiding the binarized encoder to focus on components with larger information loss.
Importantly, the auxiliary objective is used only during training and removed after convergence, introducing no extra cost during inference. Consequently, BiLoc preserves the efficiency advantages of 1-bit networks while significantly improving representation quality. As illustrated in Fig.~\ref{fig: overview}, extensive experiments on large-scale outdoor LiDAR datasets demonstrate that BiLoc establishes a new benchmark for BNN-based LiDAR localization. The main contributions are summarized as follows:

\begin{itemize}
\item[$\bullet$] We propose the {BiLoc} method, the first BNN framework tailored for outdoor localization. By constraining both weights and activations to 1-bit, BiLoc substantially reduces computational and memory costs while maintaining competitive localization accuracy.

\item[$\bullet$] We identify that the primary challenge of applying BNNs to LiDAR localization lies in their limited representational capacity and optimization difficulty. To address this, we introduce a training-only auxiliary objective that promotes the preservation of task-relevant information at the representation level without adding inference overhead.

\item[$\bullet$] We conduct extensive experiments on large-scale outdoor LiDAR datasets to validate the effectiveness of the proposed framework. BiLoc consistently outperforms existing binarization-based localization methods and establishes a new state-of-the-art (SOTA) in this field. In particular, on the Oxford Radar RobotCar dataset, BiLoc reduces the averaged mean position error by 10.11\% and the averaged mean orientation error by 11.16\% compared with the strongest binarized baseline.
\end{itemize}

%10.11%error_t, 11.16%error_q

\section{Related Work}
\label{sec:related}

\textbf{Learning-based localization} can be broadly divided into structure-based Scene Coordinate Regression (SCR) and regression-based Absolute Pose Regression (APR). SCR estimates the global pose by performing 3D matching between LiDAR point clouds and the world coordinate system. Most mainstream methods~\cite{uy2018pointnetvlad, wang2019deep, komorowski2021minkloc3d, chen2023sc} rely on learned descriptors and require storing point cloud descriptors for pose estimation.
SGLoc~\cite{li2023sgloc} is the first method to regress correspondences with a neural network and estimate pose with RANSAC, inspiring follow-up works~\cite{yang2024lisa, li2025lightloc, yang2025raloc}. However, since only the correspondence stage is trainable, its accuracy is significantly affected by the RANSAC module.
In contrast, APR regresses the 6-DoF pose end-to-end~\cite{chen2022dfnet, kendall2017geometric, kendall2015posenet, moreau2022lens, shavit2023coarse, wang2020atloc}, avoiding point cloud registration and thus being more efficient at inference. 
Since LiDAR is robust to illumination changes, LiDAR-based APR methods have achieved strong performance in large-scale outdoor environments. PointLoc first introduces APR for LiDAR localization, while~\cite{yu2022lidar} explores memory-efficient regression schemes with four models. HypLiLoc fuses multi-modal scene features in hyperbolic and Euclidean spaces. DiffLoc further adopts vision foundation models~\cite{oquab2023dinov2} for feature extraction and uses diffusion models~\cite{wang2023posediffusion} for iterative pose regression.

\noindent\textbf{Auxiliary objective} was initially introduced~\cite{szegedy2015going} to address the difficulty of training deep neural networks. These objectives are only active during training and are removed after training, so they do not introduce extra computation during inference. In recent years, auxiliary objectives have been widely used in local learning~\cite{belilovsky2020decoupled, duan2022training, wang2025infopro}, where gradient-isolated submodules are trained with module-specific losses to avoid cross-module backpropagation.
Beyond local learning, auxiliary objectives have also been linked to knowledge distillation (KD). Zhang et al.~\cite{zhang2019your} employ multi-exit self-distillation, where intermediate exits act as auxiliary supervision, and Deng et al.~\cite{deng2023surrogate} further formalize KD as a training-time auxiliary objective to provide more reliable gradients for spiking neural networks~\cite{deng2023surrogate}.
In BNNs, auxiliary designs have been explored for object detection~\cite{xu2024learning} and change detection~\cite{yin2025information}, demonstrating their effectiveness in improving optimization and performance.

\noindent\textbf{Binary neural networks (BNNs)} represent the extreme of quantization that reduces both weights and activations to 1-bit, substantially lowering computation and memory but often incurring notable accuracy drops compared with real-valued models. To mitigate this problem, many approaches have been proposed. For instance, scaling factors are applied to weights and activations to alleviate quantization error~\cite{rastegari2016xnor}; real-valued identity shortcuts are introduced to enhance representation power~\cite{birealnet}; and learnable thresholds together with RPReLU activations are adopted to better handle different data distributions~\cite{reactnet}.
For Vision Transformers (ViTs), directly applying binarization often leads to severe performance degradation~\cite{gao2025bhvit}, motivating a series of dedicated designs~\cite{bivit, bi-vit, yin2024si}. Despite this progress, most methods are validated primarily on classification, while their effectiveness on downstream tasks~\cite{BinaryDADnet, xu2022ida, chen2024binarized, pathfinder}, especially localization, remains underexplored.
To bridge this gap, we present a training optimization framework with an auxiliary objective tailored for 1-bit LiDAR localization networks, improving the performance without any additional inference cost.

\section{Preliminaries}

\subsection{Binary Neural Networks}
For the $i$-th layer, let $W_i$ and $A_i$ be the real-valued weights and activations. Binary neural networks quantize both to 1-bit, replacing floating-point multiplications with efficient bitwise operations (XNOR and PopCount).

\begin{figure}[t]
\centering
\includegraphics[width=3.7in, keepaspectratio]{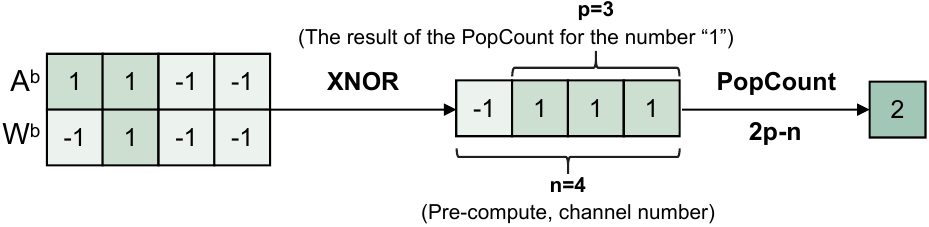}
\caption{The inner product of two binary vectors can be efficiently computed using XNOR and PopCount. After the XNOR operation, let $p$ be the number of matched bits (PopCount result) and $u$ be the number of unmatched bits, where $n = p + u$ is the vector length. Since $u = n - p$, the dot product can be expressed as $p - (n - p)$.
}
\label{fig: xnorpopcount}
\end{figure}

As the fundamental module in 1-bit ViTs, the binary linear layer ($\Lambda(\cdot)$) can be expressed as $A_{i+1}^{b}=\Lambda \left( A_i, W_i \right) =\varsigma \odot \left( A_{i}^{b}\circledast W_{i}^{b} \right) $, 
where $A_i^{b}$ and $W_i^{b}$ are binarized activations and weights, $\varsigma$ is a channel-wise scaling factor, $\circledast$ denotes the bitwise operation based on XNOR and PopCount, as shown in Fig.~\ref{fig: xnorpopcount}. $\odot$ is element-wise multiplication.
To obtain binarized activations $A^{b}$, we apply the sign function in the forward pass and use a piecewise polynomial function in the backward pass, as follows:
\begin{equation}
\resizebox{0.93\linewidth}{!}{$
\begin{aligned}
\label{eq: bi_act}
\mathrm{Forward:}~\mathrm{A}^{\mathrm{b}}&=sign\left( \frac{\mathrm{A}-\mathrm{\beta}}{\mathrm{\alpha}} \right) ,
\\
\mathrm{Backward:}~\frac{\partial L}{\partial \mathrm{A}}=\frac{\partial L}{\partial \mathrm{A}^{\mathrm{b}}}\odot \frac{\partial \mathrm{A}^{\mathrm{b}}}{\partial \mathrm{A}}&=
\left\{ \begin{matrix}
	\frac{\partial L}{\partial \mathrm{A}^{\mathrm{b}}}\odot \left( 2+2\left( \frac{\mathrm{A}-\mathrm{\beta}}{\mathrm{\alpha}} \right) \right),&		\mathrm{\beta}-\mathrm{\alpha}\leqslant \mathrm{A}<\mathrm{\beta}\\
	\frac{\partial L}{\partial \mathrm{A}^{\mathrm{b}}}\odot \left( 2-2\left( \frac{\mathrm{A}-\mathrm{\beta}}{\mathrm{\alpha}} \right) \right),&		\mathrm{\beta}\leqslant \mathrm{A}<\mathrm{\beta}+\mathrm{\alpha}\\
	0,&		\text{otherwise}\\
\end{matrix} \right., 
\end{aligned}
$}
\end{equation}
where $\alpha$ and $\beta$ are learnable scaling and bias terms, respectively. Here $L$ is the loss function.
The binarized weight $W^{b}$ commonly computed as follows:
\begin{equation}
\resizebox{\linewidth}{!}{$
\begin{aligned}
\label{eq: bi_weight}
\mathrm{Forward}:\mathrm{W}^{\mathrm{b}}=\mathrm{\rho}\left( abs\left( \mathrm{W} \right) \right) \odot sign\left( \mathrm{W} \right) ,  \mathrm{Backward}:\frac{\partial L}{\partial \mathrm{W}}=\mathrm{\rho}\left( abs\left( \mathrm{W} \right) \right) \odot \frac{\partial L}{\partial \mathrm{W}^{\mathrm{b}}}\odot \mathrm{\mu},
\end{aligned}
$}
\end{equation}
where $\rho(\cdot)$ denotes the average function used to compute the per-channel scaling factor of the weights. $\mu$ is a mask tensor with the same shape as $W$. An entry in the mask is set to one if the corresponding value in $W$ lies within the interval $[-1,1]$.

\subsection{Information-bottleneck Theory}

Information-bottleneck (IB) theory provides a unified framework for balancing data compression and task-relevant information preservation. The mutual information (MI) is a measure of the shared information between two random variables $X$ and $Y$, defined as (for discrete variables)
\begin{equation}
\resizebox{0.93\linewidth}{!}{$
\begin{aligned}
\label{eq: MI}
I\left( X,Y \right) =\underset{x}{\sum{}}\underset{y}{\sum{}}p\left( x,y \right) \log \frac{p\left( x,y \right)}{p\left( x \right) p\left( y \right)}
=\underset{x}{\sum{}}p\left( x \right) \underset{y\,}{\sum{}}p\left( y|x \right) \,\,\log \left( \frac{p\left( y|x \right)}{p\left( y \right)} \right).
\end{aligned}
$}
\end{equation}
The IB theory can be viewed as a special case of rate-distortion (RD) theory~\cite{Berger1975, Gibson2025}, which trades off lossy compression and distortion. RD seeks a compact representation $T$ of a source $X$ via $p(t|x)$ by minimizing $I(X,T)$ subject to an expected distortion constraint $E[d(X,T)] \le \epsilon^*$. Here, $d(x,t)$ measures the deviation between $X$ and $T$.
A key limitation of RD is that both the distortion function $d(x,t)$ and the threshold $\epsilon^*$ must be specified a priori, which can be difficult without prior knowledge~\cite{tishby2000information}. IB addresses this by introducing task supervision $Y$ and a trade-off parameter $\sigma$, compressing $X$ into $T$ while preserving task-relevant information, formulated as
\begin{equation}
\begin{aligned}
\label{eq: originIB}
 \textbf{IB}(p(t|x)) = \min I(X, T) - \sigma I(T, Y),
\end{aligned}
\end{equation}
where $I(X, T)$ measures the compression of $X$ into $T$, and $I(T, Y)$ quantifies the retention of task-relevant information.  The parameter $\sigma$ controls the trade-off, and $p(t|x)$ defines the mapping from $x$ to $t$.
Previous studies~\cite{shwartz2017opening, zhang2022quantifying} have shown that deep networks can learn efficient representations related to minimal sufficient statistics from the IB perspective.

\section{Methodology}
\subsection{Problem Formulation}

Tishby et al.~\cite{shwartz2017opening} suggest that the hierarchy of a DNN can be viewed as a (continuous) Markov chain that progressively refines data representations. In LiDAR localization, intermediate encoder blocks produce latent features $Z=f(X)$ from an input point cloud frame $X$, retaining task-relevant information while discarding redundancy. As illustrated in Fig.~\ref{fig: teaser}a, regions critical to performance, where perturbations cause noticeable degradation, tend to shrink with increasing depth. A regressor $g(Z)$ then predicts the global pose $Y$, yielding the mapping $X \xrightarrow{f(X)} Z \xrightarrow{g(Z)} Y$.
Under the IB principle, $Z$ is encouraged to be a minimal sufficient statistic of $Y$. Following Eq.~\ref{eq: originIB}, this corresponds to maximizing $I(Z, Y)$ while minimizing $I(X, Z)$, which can be formulated by
\begin{equation}
\label{eq:IB}
\mathbf{IB}(X,Y,Z)=\min_{\theta_f,\theta_g}\; I(X,Z)-\sigma I(Z,Y),
\end{equation}
where $\theta_f$ and $\theta_g$ are the parameters of the encoder and regressor, respectively, and $I(\cdot,\cdot)$ denotes mutual information. $I(X,Z)$ encourages compressing $X$ into $Z$ by suppressing task-irrelevant redundancy, while $I(Z,Y)$ promotes preserving task-relevant information for predicting $Y$. In LiDAR localization, maximizing $I(Z, Y)$ can be instantiated by the standard localization loss $L_{\text{localization}}$.

However, Fig.~\ref{fig: teaser}b shows that BNNs achieve much lower localization accuracy than real-valued networks. Excessive information discard yields a very low $I(X, Z)$, leaving little room for further optimization; meanwhile, important task-related information is poorly preserved, making $I(Z, Y)$ hard to maximize. This effect is partly illustrated on the right part of Fig.~\ref{fig: teaser}a.
This gap stems from the binarization mechanism in Eq.~\ref{eq: bi_act} and Eq.~\ref{eq: bi_weight}: the forward pass relies on the sign function for both weights and activations, limiting representational capacity. In the backward pass, the sign function is approximated by the straight-through estimator (STE)~\cite{Bengio2013STE} and variants, whose gradients deviate from true gradients and introduce cumulative errors. This causes the intrinsic gradient mismatch problem of BNNs, leading to unstable training and degraded performance.

\subsection{Auxiliary Objective}
\textbf{Auxiliary objective Design.}
To alleviate the above issues, we introduce a feature-distillation-based auxiliary objective for training BNNs in LiDAR localization. Specifically, we train a real-valued teacher with a strong vision foundation encoder~\cite{oquab2023dinov2} on the localization task to obtain a reference model. We then use its features as a reference distribution to guide the training of the binarized encoder. Following~\cite{shen2021s2}, we adopt the efficient offline distillation scheme as the basis of the auxiliary objective. 
We view the real-valued features as an approximation to the optimal sufficient statistic, providing an approximate upper bound for $I(X, Z)$ in Eq.~\ref{eq:IB}. Based on this, BNN training can be reformulated as:
\begin{equation}
    \begin{aligned}
    \label{eq: newIB}
\min_{\theta _f, \theta _g} I(X,Z)-\sigma _1I(Z,Y)-\sigma _2\mathbf{\varphi},
    \end{aligned}
\end{equation}
where $\sigma_1$ and $\sigma_2$ are trade-off coefficients balancing information compression and task-relevant information retention. We define a structural operator $S(\cdot)$ to capture the inter-sample relational structure of latent features, which are further used to construct a cost matrix $C$. Here, $\varphi$ denotes the auxiliary objective:
\begin{equation}
\begin{aligned}
\label{eq:aux}
\mathbf{\varphi} =I(Z_{diff},Z^*)+I(S(Z),S(Z^*) \mid \mathbf{p}),
\end{aligned}
\end{equation}
where $Z_{\mathrm{diff}}$ denotes the importance-weighted latent feature by a soft-mask. $Z^*$ denotes the latent feature of the offline real-valued model, which is treated as an approximation of the optimal sufficient statistic.
% $I(Z_{\mathrm{diff}}, Z^*)$ represents a soft-masked feature distillation term for the BNN encoder to learn the latent features of offline real-valued counterparts, which strengthens learning on important regions while suppressing variations in irrelevant and redundant regions. 
$I(Z_{\mathrm{diff}}, Z^*)$ is a soft-masked feature distillation term for the BNN encoder to learn the latent features of offline real-valued counterparts, which emphasizes important regions while suppressing variations in irrelevant regions. 
$I(S(Z), S(Z^*) \mid \mathbf{p})$ is a task-specific conditional mutual information term. With pose as side information, it relaxes the strict one-to-one correspondence between samples in $Z$ and $Z^*$, denoted as $z$ and $z^{*}$, encouraging each sample in $Z$ to align not only with its corresponding sample in $Z^*$ but also with its neighboring samples, which have similar poses.

\noindent\textbf{Auxiliary Objective Loss.}
Combining Eq.~\ref{eq: newIB} and Eq.~\ref{eq:aux}, we obtain the final objective for BiLoc training. We instantiate the auxiliary objective in Eq.~\ref{eq:aux} with a feature distillation loss using a soft mask and a structure distillation loss. For $I(Z_{\text{diff}}, Z^*)$, the auxiliary objective encourages the BNN encoder to learn task-critical components from the offline teacher features while suppressing redundant information. In BiLoc, these components correspond to different feature channels. Since obtaining a knowledge-points mask~\cite{zhang2022quantifying} is extremely time-consuming during training, we instead compute the channel-wise Mahalanobis distance between real-valued features and their 1-bit counterparts, and apply a sigmoid function to obtain a soft mask:
\begin{equation}
\begin{aligned}
\label{eq:mask}
\mathbf{m}=\varsigma \!\left( d_{M}^{2}\!\left( Z_{c}^{*},\,Z_c;\,\mathbf{\Sigma }_c \right) \right) ,\quad d_{M}^{2}(\mathbf{a},\mathbf{b};\mathbf{\Sigma })=(\mathbf{a}-\mathbf{b})^{\top}\mathbf{\Sigma }^{-1}(\mathbf{a}-\mathbf{b}),
\end{aligned}
\end{equation}
where $\mathbf{m}$ denotes the mask, $c$ denotes the channel, $\varsigma(\cdot)$ denotes the sigmoid function, and $d_{M}^{2}(\cdots)$ denotes the Mahalanobis distance. In practice, all channels share a learnable variance scalar for computational efficiency. We apply this mask to reweight features for distillation and use an $L_{1}$ loss as the distillation objective, which approximately minimizes $-I(Z_{\text{diff}}, Z^*)$: $||\mathbf{m}\odot Z-Z^*||_1$.

For $I(S(Z),S(Z^*) \mid \mathbf{p})$, we instantiate this term based on the optimal transport principle~\cite{villani2008optimal}.
Specifically, following prior work~\cite{chen2021wasserstein}, we model the structural alignment between teacher and student features as a primal optimal transport problem. It can be interpreted as finding an optimal matching that transfers probability mass from $S(Z)$ to $S(Z^*)$ when only limited training samples are available. 
In practice, we follow previous studies~\cite{sarlin2020superglue} to use the Sinkhorn algorithm~\cite{cuturi2013sinkhorn} to obtain an efficient approximation.
The Sinkhorn algorithm 
% is a differentiable relaxation of the Hungarian algorithm~\cite{munkres1957algorithms}. It 
performs iterative normalization over rows and columns, which is equivalent to applying softmax operations along both dimensions. We denote the Sinkhorn operator as $ \lVert \cdot,; C \rVert_s $, where $C \in \mathbb{R}^{N \times M}$ is the cost matrix that measures the discrepancy between two feature sets. Specifically, the matching cost between the $i$-th teacher sample $Z_i^{*}$ and the $j$-th student sample $Z_j$ is defined as 
% $C_{ij}=\bigl( 1-\cos(Z_{i}^{*},Z_j) \bigr) + d_{\mathrm{pose}}(Z_{i}^{*},Z_j),$
\begin{equation}
\begin{aligned}
\label{eq:cost_matrix}
C_{ij}=\bigl( 1-\cos(Z_{i}^{*},Z_j) \bigr) + d_{\mathrm{pose}}(Z_{i}^{*},Z_j),
\end{aligned}
\end{equation}
where $i$ and $j$ denote the sample indices in the teacher and student batches, respectively. Here, the first term is the cosine cost that measures the feature discrepancy between $Z_i^{*}$ and $Z_j$. The second term is a pose-aware cost tailored for LiDAR localization, which encodes the geometric inconsistency between the corresponding poses, which is defined as:
\begin{equation}
\begin{aligned}
\label{eq:d_pose}
d_{\mathrm{pose}}(Z_{i}^{*},Z_j)=\left\| t_i-t_j \right\| _2+\,\theta^{'} (q_i,q_j), \ \
\theta^{'} (q_i,q_j)=2\mathrm{arc}\cos \bigl( |q_{i}^{\top}q_j| \bigr) ,
\end{aligned}
\end{equation}
where $t$ is the ground-truth translation vector of each sample, and $q$ is the corresponding rotation represented as a quaternion. $\theta^{'}(\cdot)$ denotes the minimal rotation angle between two poses.
Under this setting, the loss function can be described as:
\begin{equation}
\begin{aligned}
\label{eq:loss_s}
||S\left( Z \right) ,S\left( Z^* \right) ;C||_s\triangleq \mathop {\mathrm{arg}\min} \limits_{\pi \in \mathbf{\Pi }\left( S(Z),S(Z^*) \right)}\;\langle \pi ,C\rangle +\varepsilon \sum_{i,j}{\pi _{ij}\log\mathrm{(}\pi _{ij}),}
\end{aligned}
\end{equation}
where $\pi$ denotes the discrete joint distribution of $S(Z)$ and $S(Z^*)$, which is also known as the transport plan. It satisfies the constraint: $\mathbf{\Pi} \left( S(Z), S(Z^*) \right) =\left\{ \pi \,\,| \pi \textbf{1}=S(Z), \pi ^{\mathrm{\top}}\textbf{1}=S(Z^*) \right\}$, where $\textbf{1}$ denotes an all-ones vector such that the matrix–vector multiplication enforces the marginal (row and column sum) constraints.
The term $\langle \pi, C \rangle = \sum_{i,j} \pi_{i,j} C_{i,j}$ denotes the total matching cost, which is computed in practice using the Frobenius dot product. The last term is a convex regularization term in the Sinkhorn algorithm. We use the optimal $\pi$ as a proxy loss to approximately minimize $-I(S(Z), S(Z^*) \mid \mathbf{p})$.

To clarify the formula’s meaning, we repositioned the hyperparameters in the following equation. Accordingly, we optimize the 1-bit LiDAR localization mode based on the total loss as:
\begin{equation}
\begin{aligned}
\label{eq:loss_total}
\mathcal{L}_{total}
=\underbrace{\mathcal{L}_{\text{localization}}}_{I(X;Z)-\sigma I(Z;Y)}
+\lambda_1\underbrace{\lVert \mathbf{m}\odot Z-Z^*\rVert_1}_{-I(Z_{\text{diff}};Z^*)}
+\lambda_2\underbrace{\lVert S(Z),S(Z^*);C\rVert_{s}}_{-I(S(Z),S(Z^*) \mid \mathbf{p})}.
\end{aligned}
\end{equation}
 
\noindent\textbf{Theoretical Analysis.}
We consider a binarized network decomposed into an encoder $f_e(\cdot;\theta_e)$ and a task head
$f_c(\cdot;\theta_c)$. Given an input $X$ and label $y$, the prediction is
$y_c=f_c(f_e(X;\theta_e);\theta_c)$.
Due to the non-differentiability of the sign function, training relies on the Straight-Through
Estimator (STE), which introduces systematic gradient mismatch during backpropagation through binarized layers, as shown in Eq.~\ref{eq: bi_act} and~\ref{eq: bi_weight}.
Let $G$ denote an ``accurate'' reference gradient for the encoder, and let $\upsilon$ represent the
accumulated STE-induced error. The encoder-side gradient propagated from the task head can be modeled as $\mathrm{STE}\!\left(\frac{\partial y_c}{\partial \theta_e}\right)=G+\upsilon$.
For baseline training using only the task loss $\mathcal{L}_{\text{localization}}$, which denoted as $\mathcal{L}_{\text{loc}}$ for simplicity, the encoder receives $\frac{\partial \mathcal{L}_{\text{loc}}}{\partial \theta_e}=\frac{\partial \mathcal{L}_{\text{loc}}}{\partial y_c}(G+\upsilon),$
% \begin{equation}
% \frac{\partial \mathcal{L}_{\text{loc}}}{\partial \theta_e}
% =
% \frac{\partial \mathcal{L}_{\text{loc}}}{\partial y_c}(G+\upsilon),
% \label{eq:grad_base}
% \end{equation}
and the relative gradient mismatch can be characterized by
$K_{\text{base}}=\|\upsilon\|/\|G\|$.

As mentioned in earlier chapters, BiLoc introduces auxiliary objectives defined directly on the encoder representation to provide
additional gradient routes independent of the task head, forming the total objective shown in Eq.~\ref{eq:loss_total}.
Let
\begin{equation}
\varPhi_{\text{rep}}
=
\frac{\partial Z}{\partial \theta_e}
\frac{\partial \mathcal{L}_{\text{rep}}}{\partial Z},
\quad
\varPhi_{\text{struct}}
=
\frac{\partial Z}{\partial \theta_e}
\frac{\partial \mathcal{L}_{\text{struct}}}{\partial Z}
\label{eq:S_defs}
\end{equation}
denote the encoder gradients contributed by representation distillation and struc-ture-aware supervision, respectively, where $\mathcal{L}_{\text{rep}}$ represents $-I(Z_{\text{diff}};Z^*)$, and $\mathcal{L}_{\text{struct}}$ represents $-I(S(Z), S(Z^*) \mid \mathbf{p})$.
These gradients bypass the task head and avoid accumulating head-side STE mismatch, providing more stable auxiliary signals for updating $\theta_e$.
The overall encoder gradient becomes
\begin{equation}
\frac{\partial \mathcal{L} _{\mathrm{total}}}{\partial \theta _e}=\frac{\partial \mathcal{L} _{\mathrm{loc}}}{\partial y_c}(G+\upsilon )+\lambda _{1}\varPhi _{\mathrm{rep}}+\lambda _{2}\varPhi _{\mathrm{struct}}.
\label{eq:grad_total}
\end{equation}
Accordingly, we define the relative mismatch proxy as
\begin{equation}
K_{\mathrm{BiLoc}}=\parallel \upsilon \parallel /\parallel G+\lambda _1\varPhi _{\mathrm{rep}}+\lambda _2\varPhi _{\mathrm{struct}}\parallel .
\label{eq:K_biloc}
\end{equation}
% Provided that the auxiliary gradients are not strongly adversarial to $G$, we have
% $\|G+\lambda_{1} \varPhi_{\text{rep}}+\lambda_{2} \varPhi_{\text{struct}}\|\ge \|G\|$, yielding
% $K_{\text{BiLoc}}\le K_{\text{base}}$.
By the vector squared sum identity $\|a+b\|^2 = \|a\|^2 + \|b\|^2 + 2\langle a,b\rangle$, we have
\begin{equation}
\resizebox{1.0\linewidth}{!}{$
\|G+\lambda _1\Phi _{\mathrm{rep}}+\lambda _2\Phi _{\mathrm{struct}}\|^2
=
\|G\|^2
+
\|\lambda _1\Phi _{\mathrm{rep}}+\lambda _2\Phi _{\mathrm{struct}}\|^2
+
2\langle G,\lambda _1\Phi _{\mathrm{rep}}+\lambda _2\Phi _{\mathrm{struct}}\rangle.
$}
\end{equation}
Therefore, a sufficient condition for $\|G+\lambda_{1}\Phi_{\text{rep}}+\lambda_{2}\Phi_{\text{struct}}\|
\ge
\|G\|$
is
\begin{equation}
\Big\langle 
G,\lambda_{1}\Phi_{\text{rep}}+\lambda_{2}\Phi_{\text{struct}}
\Big\rangle
\ge
-\frac{1}{2}
\|\lambda_{1}\Phi_{\text{rep}}+\lambda_{2}\Phi_{\text{struct}}\|^2.
\end{equation}
In particular, when the auxiliary gradients are not adversarial to the task gradient in the sense that $\Big\langle 
G,\lambda_{1}\Phi_{\text{rep}}+\lambda_{2}\Phi_{\text{struct}}
\Big\rangle \ge 0,$
the inequality holds directly, yielding $K_{\text{BiLoc}} \le K_{\text{base}}.$
In practice, representation-level supervision enlarges the effective gradient magnitude, while the structure-aware components reduce misaligned or noisy auxiliary gradients, making them more consistent with the task gradient. As a result, BiLoc alleviates the relative impact of STE-induced gradient mismatch in deep binarized encoders by injecting stable and task-relevant auxiliary gradients at the representation level.

\section{Experiments}
\subsection{Experimental Setup}
\textbf{Dataset and Metrics.} We evaluate proposed BiLoc for LiDAR localization on two large-scale outdoor datasets: Oxford Radar RobotCar~\cite{barnes2020oxford}, which we denote as Oxford for simplicity, and NCLT~\cite{carlevaris2016university} datasets. During the evaluation, we select the mean position and orientation error as the evaluation metrics.

\noindent\textbf{Oxford Radar RobotCar}  dataset is a widely used benchmark for urban localization. It covers about 2 $km^2$, and each trajectory is nearly 10 km long. The dataset provides multi-sensor data, including LiDAR, cameras, radar, and GPS/INS. In our work, we use only LiDAR data. Its diverse traffic conditions make it suitable for evaluating LiDAR-based localization methods. Following prior studies~\cite{li2023sgloc, yang2025raloc}, we use sequences 11-14-02-26, 14-12-05-52, 14-14-48-55, and 18-15-20-12 for training, and 15-13-06-37, 17-13-26-39, 17-14-03-00, and 18-14-14-42 for testing.

\noindent\textbf{NCLT} dataset is a campus area localization dataset collected at the Michigan North Campus. It consists of data from the LiDAR, omnidirectional camera, and GPS/INS. In our experiments, we use only LiDAR information. It includes 27 tracks, and each track is about 5.5 $km$ long, covering an area of 0.45 $km^2$. The dataset contains many challenging scenes, such as moving objects and seasonal variations, and structural changes caused by construction.

\noindent\textbf{Implementation details.}
Our method is implemented in PyTorch. The range image size and patch size are set to $[32, 512]$ and $[4, 16]$, respectively, with a batch size of 50. The input point cloud sequence consists of three frames with a stride of two. The number of denoising steps is set to 10 on Oxford and 15 on NCLT. We train the model for 200 epochs using AdamW~\cite{dosovitskiy2021image} with an initial learning rate of $1\text{e-3}$ and a cosine annealing schedule on a single RTX 5090 GPU. For the auxiliary objective, the offline teacher encoder is DINO~\cite{oquab2023dinov2} with a ViT-S/16 backbone, pre-trained on the corresponding dataset, while the rest of the network is trained from scratch. The auxiliary objective follows the same initial learning rate and warm-up schedule as the backbone, and its gradient update to the backbone stops after the 80th epoch~\cite{cho2019efficacy}.

\subsection{Comparison with SOTA BNNs}

\textbf{Baselines and comparisons.}
As the first study to explore LiDAR localization based on BNNs, we compare our proposed BiLoc with existing ViT-adapted BNNs, including ReActNet~\cite{reactnet}, BinaryViT~\cite{Le_2023_CVPR}, and BHViT~\cite{gao2025bhvit}, which have achieved state-of-the-art performance on classification tasks. We adapt ReActNet to the ViT-S/16~\cite{touvron2021training} architecture for our experiments. We re-run the official code released by the authors of these SOTA BNNs and train all models under the same settings. We adopt DiffLoc~\cite{li2024diffloc}, an SOTA APR method, as the evaluation framework and integrate BiLoc and all baseline BNNs into this framework for fair comparison. The results of real-valued networks are also included as references. Memory usage follows the mainstream approach~\cite{reactnet, Wang2020BiDetAE}: 32$\times $ real-valued kernels + 1$\times $ binary kernels. Operations (OPs) are calculated as real-valued FLOPs plus 1/64 of 1-bit multiplications, following Bi-Real-Net's protocol~\cite{birealnet}.  

Table~\ref{tab:complexity} compares the computational complexity and memory consumption of different quantization strategies and LiDAR localization frameworks.
Benefiting from a lighter backbone and 1-bit quantization, our method achieves significant improvements in both computational efficiency and memory usage compared with its real-valued version. This improvement effectively alleviates the computational bottleneck of real-valued DiffLoc and makes the framework more suitable for deployment in resource-constrained environments.
\begin{table}[t]
\centering
\caption{Comparison of inference-time model complexity for real-valued and 1-bit LiDAR localization models. 
Params are reported in millions (M), and OPs are reported in giga-operations (G).}
\label{tab:complexity}

\renewcommand{\arraystretch}{1.0}
\setlength{\tabcolsep}{5pt}

\scalebox{0.9}{
\begin{tabular}{lcc lcc}
\toprule
\multicolumn{3}{c}{Real-valued Models} 
& \multicolumn{3}{c}{1-bit Models} \\
\cmidrule(r){1-3} \cmidrule(l){4-6}
Method & Params (M) & OPs (G) 
& Method & Params (M) & OPs (G) \\
\midrule

PointLoc & 3.29 & 10.13 
& DiffLoc + ReActNet & 2.89 & 0.38 \\

STCLoc & 9.31 & 1.59 
& DiffLoc + BinaryViT & 2.83 & 0.14 \\

HypLiLoc & 52.31 & 4.89 
& DiffLoc + BHViT & 3.01 & 0.17 \\

DiffLoc  & 39.96 & 77.06 
& DiffLoc + Ours & 3.01 & 0.17 \\

\bottomrule
\end{tabular}
}
\end{table}

\noindent\textbf{Results on the Oxford dataset.}
Table~\ref{tab:oxford} evaluates the performance of the proposed method on the Oxford dataset. We report the mean position error and orientation error averaged over all test trajectories. Our method achieves an averaged mean error of $7.56\text{m}/1.91^\circ$, reducing the error of the previous SOTA BNNs method by $0.85\text{m}/0.24^\circ$, without introducing any additional computational or memory overhead. 
\begin{table}[t]  
\centering
\caption{Quantitative results of real-valued LiDAR localization models, 1-bit LiDAR localization models, and our proposed methods. The evaluation covers the mean position(\text{m}) and orientation($^\circ$) error on the Oxford dataset.}
\label{tab:oxford}

\renewcommand{\arraystretch}{1.0}
\setlength{\tabcolsep}{2pt}
\scalebox{0.90}{
\begin{tabular}{cccccccc}  
\toprule
Framework & Method & Bits  & \multicolumn{1}{c}{\begin{tabular}[c]{@{}c@{}} 15-13- \\ 06-37\end{tabular}} & \multicolumn{1}{c}{\begin{tabular}[c]{@{}c@{}} 17-13- \\ 26-39\end{tabular}} & \multicolumn{1}{c}{\begin{tabular}[c]{@{}c@{}} 17-14- \\ 03-00\end{tabular}} & \multicolumn{1}{c}{\begin{tabular}[c]{@{}c@{}} 18-14- \\ 14-42\end{tabular}} & \multicolumn{1}{c}{\begin{tabular}[c]{@{}c@{}} Average \\ $\, [\text{m}/^\circ] \,$\end{tabular}}\\
\midrule
PointLoc~\cite{wang2021pointloc}
  & Real-valued      
  & 32    
  & 12.42/2.26
  & 13.14/2.50
  & 12.91/1.92
  & 11.31/1.98
  & 12.45/2.17         \\

\midrule
PosePN~\cite{yu2022lidar}
  & Real-valued      
  & 32     
  & 14.32/3.06
  & 16.97/2.49
  & 13.48/2.60
  & 9.14/1.78
  & 13.48/2.48          \\
 
\midrule
PosePN++~\cite{yu2022lidar}
  & Real-valued      
  & 32     
  & 9.59/1.92
  & 10.66/1.92
  & 9.01/1.51
  & 8.44/1.71
  & 9.43/1.77          \\

\midrule
PoseMinkLoc~\cite{yu2022lidar}
  & Real-valued      
  & 32     
  & 11.20/2.62
  & 14.24/2.42
  & 12.35/2.46
  & 10.06/2.15
  & 11.96/2.41         \\

\midrule
PoseSOE~\cite{yu2022lidar}
  & Real-valued      
  & 32     
  & 7.59/1.94
  & 10.39/2.08
  & 9.21/2.12
  & 7.27/1.87
  & 8.62/2.00\\

\midrule
STCLoc~\cite{yu2022stcloc}
  & Real-valued      
  & 32     
  & 6.93/1.48
  & 7.55/1.23
  & 7.44/1.24
  & 6.13/1.15
  & 7.01/1.28\\

\midrule
HypLiLoc~\cite{wang2023hypliloc}
  & Real-valued      
  & 32     
  & 6.88/1.09
  & 6.79/1.29
  & 5.82/0.97
  & 3.45/0.84
  & 5.74/1.05\\

\midrule 
\multirow{5}*{DiffLoc~\cite{li2024diffloc}}
  & Real-valued      
  & 32    
  & 3.57/0.88
  & 3.65/0.68
  & 4.03/0.70
  & 2.86/0.60
  & 3.53/0.72          \\

\cdashline{2-8} 
  & ReActNet           
  & 1
  & 13.56/3.21 
  & 18.73/4.54 
  & 16.91/4.87
  & 11.42/3.00 
  & 15.16/3.91 \\
  
% \cdashline{2-6} 
  & BinaryViT           
  & 1
  & 7.94/2.44
  & 11.92/2.87
  & 8.99/2.83
  & 6.61/2.15
  & 8.87/2.57 \\
  
% \cdashline{2-6} 
  & BHViT           
  & 1
  & 7.35/2.02
  & 11.44/2.45
  & 8.97/2.43
  & 5.86/1.68
  & 8.41/2.15 \\

% \cdashline{2-6} 
  & Ours          
  & 1
  & 5.97/2.03
  & 10.96/2.18
  & 8.42/2.00 
  & 4.89/1.45
  & \textbf{7.56/1.91} \\

\bottomrule
\end{tabular}}
\end{table}

\noindent\textbf{Results on the NCLT dataset.}
Since the NCLT dataset covers various seasonal changes and spans a longer time period, it poses a greater challenge for localization compared with the Oxford dataset. As shown in Table~\ref{tab:nclt}, our method achieves an averaged mean error of $3.14\text{m}/4.36^\circ$. Compared with the previous state-of-the-art binary neural network method, our approach reduces the error by $0.42\text{m}/0.11^\circ$ while incurring no additional computational or memory cost.

\begin{table}[t]  
\centering
\caption{Quantitative results on the NCLT dataset. The evaluation covers the mean position(\text{m}) and orientation($^\circ$) error.}

\label{tab:nclt}

\renewcommand{\arraystretch}{1.0}
\setlength{\tabcolsep}{2pt}
\scalebox{0.90}{
\begin{tabular}{cccccccc}  
\toprule
Framework & Method & Bits  & \multicolumn{1}{c}{\begin{tabular}[c]{@{}c@{}} 2012- \\ 02-12\end{tabular}} & \multicolumn{1}{c}{\begin{tabular}[c]{@{}c@{}} 2012- \\ 02-19\end{tabular}} & \multicolumn{1}{c}{\begin{tabular}[c]{@{}c@{}} 2012- \\ 03-31\end{tabular}} & \multicolumn{1}{c}{\begin{tabular}[c]{@{}c@{}} 2012- \\ 05-26\end{tabular}} & \multicolumn{1}{c}{\begin{tabular}[c]{@{}c@{}} Average \\ $\, [\text{m}/^\circ] \,$\end{tabular}}\\
\midrule
PointLoc~\cite{wang2021pointloc}
  & Real-valued      
  & 32    
  & 7.23/4.88
  & 6.31/3.89
  & 6.71/4.32
  & 10.02/5.32
  & 7.57/4.60         \\

\midrule
PosePN~\cite{yu2022lidar}
  & Real-valued      
  & 32     
  & 9.45/7.47
  & 6.15/5.05
  & 5.79/5.28
  & 13.47/7.77
  & 8.72/6.39          \\
 
\midrule
PosePN++~\cite{yu2022lidar}
  & Real-valued      
  & 32     
  & 4.97/3.75
  & 3.68/2.65
  & 4.35/3.38
  & 9.59/4.49
  & 5.65/3.57          \\

\midrule
PoseMinkLoc~\cite{yu2022lidar}
  & Real-valued      
  & 32     
  & 6.24/5.03
  & 4.87/3.94
  & 4.23/4.03
  & 10.32/6.52
  & 6.42/4.88\\

\midrule
PoseSOE~\cite{yu2022lidar}
  & Real-valued      
  & 32     
  & 13.09/8.05
  & 6.16/4.51
  & 5.24/4.56
  & 12.60/7.67
  & 9.27/6.20\\

\midrule
STCLoc~\cite{yu2022stcloc}
  & Real-valued      
  & 32     
  & 4.91/4.34
  & 3.25/3.10
  & 3.75/4.04
  & 8.67/5.23
  & 5.15/4.18\\

\midrule
HypLiLoc~\cite{wang2023hypliloc}
  & Real-valued      
  & 32     
  & 1.71/3.56
  & 1.68/2.69
  & 1.52/2.90
  & 2.90/3.47
  & 1.95/3.16\\

\midrule 
\multirow{5}*{DiffLoc~\cite{li2024diffloc}}
  & Real-valued      
  & 32    
  & 0.99/2.40
  & 0.92/2.14
  & 0.98/2.27
  & 1.88/2.43
  & 1.19/2.31\\

\cdashline{2-8} 
  & ReActNet           
  & 1
  & 5.76/6.78
  & 5.05/6.07
  & 4.91/6.45
  & 10.72/7.16
  & 6.61/6.62\\
  
% \cdashline{2-6} 
  & BinaryViT           
  & 1
  & 3.68/5.89
  & 3.43/4.92
  & 3.44/5.29
  & 8.21/5.80
  & 4.69/5.48\\
  
% \cdashline{2-6} 
  & BHViT           
  & 1
  & 2.69/4.81
  & 2.28/3.90
  & 2.39/4.28
  & 6.86/4.88
  & 3.56/4.47\\

% \cdashline{2-6} 
  & Ours          
  & 1
  & 2.15/4.71
  & 1.95/3.73
  & 1.94/4.15
  & 6.52/4.87
  & \textbf{3.14/4.36}\\

\bottomrule
\end{tabular}}
\end{table}

\begin{figure}[b]
\centering
\includegraphics[width=4.7in, keepaspectratio]{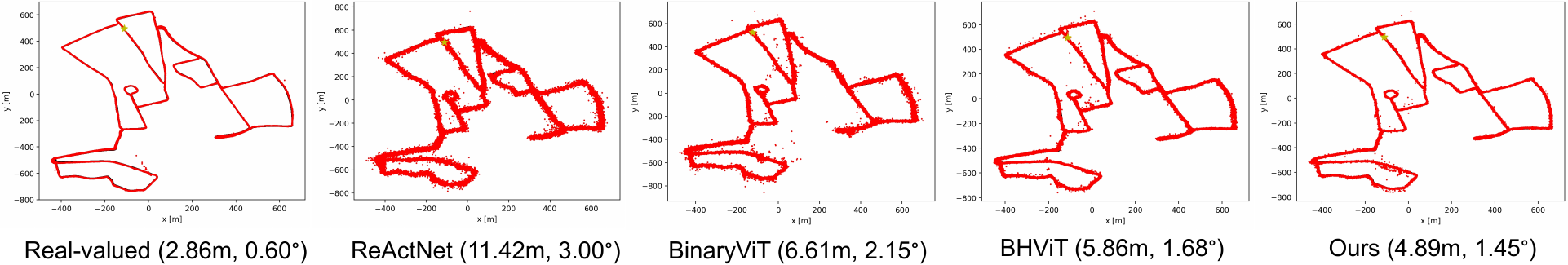}
\caption{
LiDAR localization results on the Oxford dataset (18-14-14-42 subset). Ground truth and predictions are shown in black and red, respectively, and the star marks the first frame. Each subfigure reports the mean position and orientation errors.}
\label{fig: oxfordvis}
\end{figure}

\begin{figure}[t]
\centering
\includegraphics[width=4.7in, keepaspectratio]{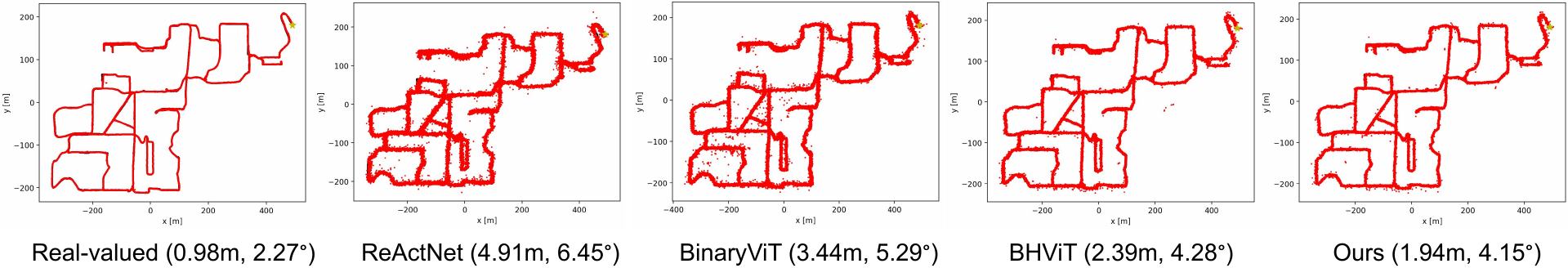}
\caption{
LiDAR localization results on the NCLT dataset (2012-03-31 subset). Ground truth and predictions are shown in black and red, respectively, and the star marks the first frame. Each subfigure reports the mean position and orientation errors.}
\label{fig: ncltvis}
\end{figure}

\noindent\textbf{Visualization.}
Fig.~\ref{fig: oxfordvis} and Fig.~\ref{fig: ncltvis} show the predicted trajectories on the 8-14-14-42 subset of Oxford and the 2012-03-31 subset of NCLT, respectively. The mean position error and orientation error are also reported. Compared with other 1-bit methods, BiLoc exhibits fewer outliers and produces trajectories that are closer to its real-valued counterpart, indicating more robust localization results.

\subsection{Ablation Studies}
\begin{figure}
\centering
\includegraphics[width=4.0in, keepaspectratio]{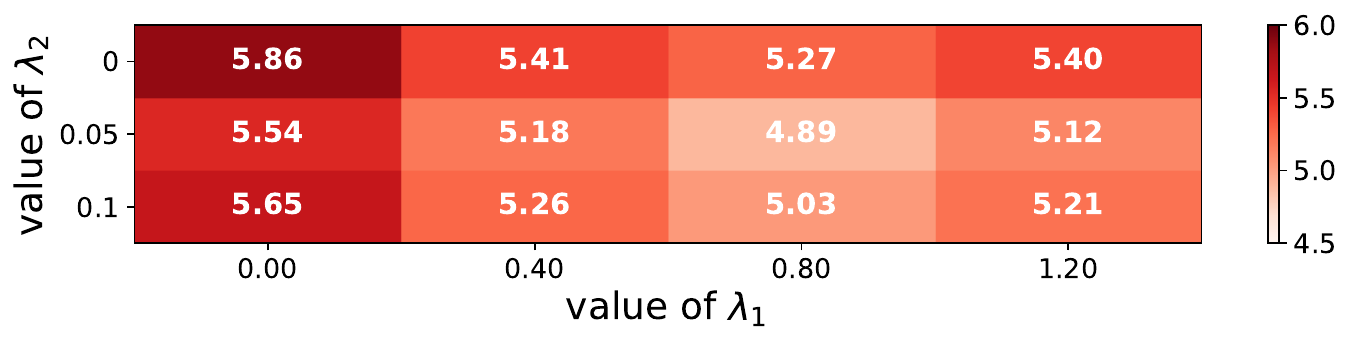}
\caption{
The hyper-parameters $\lambda_1$ and $\lambda_2$ in our BiLoc framework are selected through empirical evaluation on the 18-14-14-42 subset of the Oxford dataset.}
\label{fig: ablation}
\end{figure}

\textbf{Hyper-Parameter Selection.}
We use the 1-bit DiffLoc framework with the BHViT backbone to determine the hyperparameters $\lambda_1$ and $\lambda_2$. Figure~\ref{fig: ablation} reports the mean position error under different settings. 
For $\lambda_1$, which controls the strength of feature learning from the offline real-valued counterpart, the performance improves as $\lambda_1$ increases and reaches its optimum at $\lambda_1 = 0.80$. A larger value leads to slight degradation, suggesting that overly strong supervision may hinder stable optimization. We therefore fix $\lambda_1$ to 0.80 in subsequent experiments.
For $\lambda_2$, which encourages structural alignment between the 1-bit features and their real-valued counterparts, the best performance is achieved at $\lambda_2 = 0.05$. This indicates that moderate structural regularization is beneficial, while excessive alignment may restrict model flexibility.
Notably, the original baseline ($\lambda_1 = 0, \lambda_2 = 0$) performs worse than all variants, confirming the effectiveness and necessity of our method.

\noindent\textbf{Latency Comparison.}
We evaluate real-time latency on an RTX 5090 GPU for both binarized and real-valued models. Real-valued operations were executed with cuDNN~\cite{chetlur2014cudnn}, while binary operations are implemented using TC-BNN~\cite{li2020accelerating}. As shown in Table~\ref{tab:latency}, the 1-bit variant achieves a $2.07\times$ latency reduction compared with the real-valued baseline,  while our method adds no inference overhead. 
Due to the limited range of 1-bit operators supported by TC-BNN, the achieved inference speedup does not reach the theoretical upper bound. We expect that latency can be further reduced with the development of hardware accelerators specifically designed for binarized operations.

\begin{table}
\centering
\setlength{\tabcolsep}{4pt}
\renewcommand{\arraystretch}{1}
\caption{Latency comparison in deployment (input size: $32\times512$).}
\label{tab:latency}
\scalebox{0.9}{
\begin{tabular}{lcccc}
\toprule
Method & Bits & Params (M) & OPs (G) & Latency (ms) \\
\midrule
Real-valued DiffLoc   & 32 & 39.96 & 77.06 & 30.72 \\
1-bit DiffLoc + BHViT & 1  & 3.01  & 0.17  & 14.83 \\
1-bit DiffLoc + Ours  & 1  & 3.01  & 0.17  & 14.83 \\
\bottomrule
\end{tabular}
}
\end{table}

\section{Conclusion}

In this paper, we propose BiLoc, a fully binarized framework for resource-efficient outdoor LiDAR localization. By introducing an auxiliary objective that provides additional optimization guidance during training, our method enhances localization robustness and alleviates the representational limitations of BNNs. Extensive experiments on large-scale outdoor LiDAR datasets validate the effectiveness of BiLoc, establishing a new state of the art in efficient LiDAR localization and offering a practical solution for deployment on resource-constrained platforms.

\section{Acknowledgement}

This work was supported by Fundo para o Desenvolvimento das Ciências e da Tecnologia of Macau (FDCT) with Reference No. 0117/2024/RIB2. And we thank the high-performance computing center at Eastern Institute of Technology and Ningbo Institute of Digital Twin for providing the computing resources.

% ---- Bibliography ----
%
% BibTeX users should specify bibliography style 'splncs04'.
% References will then be sorted and formatted in the correct style.
%
\bibliographystyle{splncs04}
\bibliography{main}

@String(CVPR  = {IEEE Conf. Comput. Vis. Pattern Recog.})

@String(AAAI  = {AAAI})

@String(CVPR  = {CVPR})

@String(CVPR= {IEEE Conf. Comput. Vis. Pattern Recog.})

@String(AAAI = {AAAI})

@inproceedings{sarlin2020superglue,
  title={Superglue: Learning feature matching with graph neural networks},
  author={Sarlin, Paul-Edouard and DeTone, Daniel and Malisiewicz, Tomasz and Rabinovich, Andrew},
  booktitle={Proceedings of the IEEE/CVF conference on computer vision and pattern recognition},
  pages={4938--4947},
  year={2020}
}

@inproceedings{gao2025bhvit,
  title={BHViT: Binarized Hybrid Vision Transformer},
  author={Gao, Tian and Zhang, Yu and Zhang, Zhiyuan and Liu, Huajun and Yin, Kaijie and Xu, Chengzhong and Kong, Hui},
  booktitle={Proceedings of the IEEE/CVF Conference on Computer Vision and Pattern Recognition},
  pages={3563--3572},
  year={2025}
}

@INPROCEEDINGS{BinaryDADnet,
  author={Frickenstein, Alexander and Vemparala, Manoj-Rohit and Mayr, Jakob and Nagaraja, Naveen-Shankar and Unger, Christian and Tombari, Federico and Stechele, Walter},
  booktitle={IEEE International Conference on Robotics and Automation}, 
  title={Binary DAD-Net: Binarized Driveable Area Detection Network for Autonomous Driving}, 
  year={2020},
  volume={},
  pages={2295-2301}
}

@article{bnn,
  title={Binarized neural networks},
  author={Hubara, Itay and Courbariaux, Matthieu and Soudry, Daniel and El-Yaniv, Ran and Bengio, Yoshua},
  journal={Advances in neural information processing systems},
  volume={29},
  year={2016}
}

@inproceedings{rastegari2016xnor,
  title={Xnor-net: Imagenet classification using binary convolutional neural networks},
  author={Rastegari, Mohammad and Ordonez, Vicente and Redmon, Joseph and Farhadi, Ali},
  booktitle={European conference on computer vision},
  pages={525--542},
  year={2016},
  organization={Springer}
}

@inproceedings{birealnet,
  title={Bi-real net: Enhancing the performance of 1-bit cnns with improved representational capability and advanced training algorithm},
  author={Liu, Zechun and Wu, Baoyuan and Luo, Wenhan and Yang, Xin and Liu, Wei and Cheng, Kwang-Ting},
  booktitle={Proceedings of the European conference on computer vision},
  pages={722--737},
  year={2018}
}

@inproceedings{reactnet,
  title={Reactnet: Towards precise binary neural network with generalized activation functions},
  author={Liu, Zechun and Shen, Zhiqiang and Savvides, Marios and Cheng, Kwang-Ting},
  booktitle={European conference on computer vision},
  pages={143--159},
  year={2020},
  organization={Springer}
}

@inproceedings{bi-vit,
  title={Bi-ViT: Pushing the Limit of Vision Transformer Quantization},
  author={Li, Yanjing and Xu, Sheng and Lin, Mingbao and Cao, Xianbin and Liu, Chuanjian and Sun, Xiao and Zhang, Baochang},
  booktitle={AAAI Conference on Artificial Intelligence},
  volume={38},
  number={4},
  pages={3243--3251},
  year={2024}
}

@inproceedings{bivit,
  title={Bivit: Extremely compressed binary vision transformers},
  author={He, Yefei and Lou, Zhenyu and Zhang, Luoming and Liu, Jing and Wu, Weijia and Zhou, Hong and Zhuang, Bohan},
  booktitle={IEEE/CVF International Conference on Computer Vision},
  pages={5651--5663},
  year={2023}
}

@article{Wang2020BiDetAE,
  title={BiDet: An Efficient Binarized Object Detector},
  author={Ziwei Wang and Ziyi Wu and Jiwen Lu and Jie Zhou},
  journal={Proceedings of the IEEE/CVF conference on computer vision and pattern recognition},
  year={2020},
  pages={2046-2055}
}

@inproceedings{xu2024learning,
  title={Learning 1-bit tiny object detector with discriminative feature refinement},
  author={Xu, Sheng and Wang, Mingze and Li, Yanjing and Lin, Mingbao and Zhang, Baochang and Doermann, David and Sun, Xiao},
  booktitle={Forty-first International Conference on Machine Learning},
  year={2024}
}

@inproceedings{deng2023surrogate,
  title={Surrogate module learning: Reduce the gradient error accumulation in training spiking neural networks},
  author={Deng, Shikuang and Lin, Hao and Li, Yuhang and Gu, Shi},
  booktitle={International Conference on Machine Learning},
  pages={7645--7657},
  year={2023},
  organization={PMLR}
}

@inproceedings{zhang2019your,
  title={Be your own teacher: Improve the performance of convolutional neural networks via self distillation},
  author={Zhang, Linfeng and Song, Jiebo and Gao, Anni and Chen, Jingwei and Bao, Chenglong and Ma, Kaisheng},
  booktitle={Proceedings of the IEEE/CVF international conference on computer vision},
  pages={3713--3722},
  year={2019}
}

@article{wang2025infopro,
  title={Infopro: Locally supervised deep learning by maximizing information propagation},
  author={Wang, Yulin and Ni, Zanlin and Pu, Yifan and Zhou, Cai and Ying, Jixuan and Song, Shiji and Huang, Gao},
  journal={International Journal of Computer Vision},
  volume={133},
  number={5},
  pages={2752--2782},
  year={2025},
  publisher={Springer}
}

@inproceedings{sandler2018mobilenetv2,
  title={Mobilenetv2: Inverted residuals and linear bottlenecks},
  author={Sandler, Mark and Howard, Andrew and Zhu, Menglong and Zhmoginov, Andrey and Chen, Liang-Chieh},
  booktitle={Proceedings of the IEEE/CVF Conference on Computer Vision and Pattern Recognition},
  pages={4510--4520},
  year={2018}
}

@article{tishby2000information,
  title={The information bottleneck method},
  author={Tishby, Naftali and Pereira, Fernando C and Bialek, William},
  journal={arXiv preprint physics/0004057},
  year={2000}
}

@inproceedings{tishby2015deep,
  title={Deep learning and the information bottleneck principle},
  author={Tishby, Naftali and Zaslavsky, Noga},
  booktitle={ IEEE Information Theory Workshop},
  pages={1--5},
  year={2015}
}

@article{shwartz2017opening,
  title={Opening the black box of deep neural networks via information},
  author={Shwartz-Ziv, Ravid and Tishby, Naftali},
  journal={arXiv preprint arXiv:1703.00810},
  year={2017}
}

@article{Bengio2013STE,
  title={Estimating or Propagating Gradients Through Stochastic Neurons for Conditional Computation},
  author={Yoshua Bengio and Nicholas L{\'e}onard and Aaron C. Courville},
  journal={ArXiv},
  year={2013},
  volume={abs/1308.3432}
}

@inproceedings{qin2020forward,
  title={Forward and backward information retention for accurate binary neural networks},
  author={Qin, Haotong and Gong, Ruihao and Liu, Xianglong and Shen, Mingzhu and Wei, Ziran and Yu, Fengwei and Song, Jingkuan},
  booktitle={Proceedings of the IEEE/CVF conference on computer vision and pattern recognition},
  pages={2250--2259},
  year={2020}
}

@Inbook{Gibson2025,
author="Gibson, Jerry D.",
title="Rate Distortion Theory",
bookTitle="Information Theoretic Principles for Agent Learning",
year="2025",
publisher="Springer Nature Switzerland",
address="Cham",
pages="77--92",
isbn="978-3-031-65388-9"
}

@Inbook{Berger1975,
author="Berger, Toby",
title="Rate Distortion Theory and Data Compression",
bookTitle="Advances in Source Coding",
year="1975",
publisher="Springer Vienna",
address="Vienna",
pages="1--39",
isbn="978-3-7091-2928-9"
}

@article{Yin2023ASO,
    author = {Huan Yin and Xuecheng Xu and Shan Lu and Xieyuanli Chen and Rong Xiong and Shaojie Shen and C. Stachniss and Yue Wang},
    journal = {International Journal of Computer Vision},
    pages = {3139 - 3171},
    title = {A Survey on Global LiDAR Localization: Challenges, Advances and Open Problems},
    volume = {132},
    year = {2024}
}

@article{Fischler1981Ransac,
  title={Random sample consensus: a paradigm for model fitting with applications to image analysis and automated cartography},
  author={Martin A. Fischler and Robert C. Bolles},
  journal={Commun. ACM},
  year={1981},
  volume={24},
  pages={381-395}
}

@inproceedings{li2024diffloc,
  title={Diffloc: Diffusion model for outdoor lidar localization},
  author={Li, Wen and Yang, Yuyang and Yu, Shangshu and Hu, Guosheng and Wen, Chenglu and Cheng, Ming and Wang, Cheng},
  booktitle={Proceedings of the IEEE/CVF Conference on Computer Vision and Pattern Recognition},
  pages={15045--15054},
  year={2024}
}

@inproceedings{wang2023hypliloc,
  title={Hypliloc: Towards effective lidar pose regression with hyperbolic fusion},
  author={Wang, Sijie and Kang, Qiyu and She, Rui and Wang, Wei and Zhao, Kai and Song, Yang and Tay, Wee Peng},
  booktitle={Proceedings of the IEEE/CVF Conference on Computer Vision and Pattern Recognition},
  pages={5176--5185},
  year={2023}
}

@inproceedings{brachmann2017dsac,
  title={Dsac-differentiable ransac for camera localization},
  author={Brachmann, Eric and Krull, Alexander and Nowozin, Sebastian and Shotton, Jamie and Michel, Frank and Gumhold, Stefan and Rother, Carsten},
  booktitle={Proceedings of the IEEE/CVF Conference on Computer Vision and Pattern Recognition},
  pages={6684--6692},
  year={2017}
}

@inproceedings{brachmann2023accelerated,
  title={Accelerated coordinate encoding: Learning to relocalize in minutes using rgb and poses},
  author={Brachmann, Eric and Cavallari, Tommaso and Prisacariu, Victor Adrian},
  booktitle={Proceedings of the IEEE/CVF Conference on Computer Vision and Pattern Recognition},
  pages={5044--5053},
  year={2023}
}

@inproceedings{li2025lightloc,
  title={LightLoc: Learning Outdoor LiDAR Localization at Light Speed},
  author={Li, Wen and Liu, Chen and Yu, Shangshu and Liu, Dunqiang and Zhou, Yin and Shen, Siqi and Wen, Chenglu and Wang, Cheng},
  booktitle={Proceedings of the IEEE/CVF Conference on Computer Vision and Pattern Recognition},
  pages={6680--6689},
  year={2025}
}

@inproceedings{yin2025information,
  title={Information-Bottleneck Driven Binary Neural Network for Change Detection},
  author={Yin, Kaijie and Zhang, Zhiyuan and Kong, Shu and Gao, Tian and Xu, Cheng-Zhong and Kong, Hui},
  booktitle={Proceedings of the IEEE/CVF International Conference on Computer Vision},
  pages={7176--7186},
  year={2025}
}

@article{shi2022improved,
  title={An improved lightweight deep neural network with knowledge distillation for local feature extraction and visual localization using images and LiDAR point clouds},
  author={Shi, Chenhui and Li, Jing and Gong, Jianhua and Yang, Banghui and Zhang, Guoyong},
  journal={ISPRS journal of photogrammetry and remote sensing},
  volume={184},
  pages={177--188},
  year={2022},
  publisher={Elsevier}
}

@article{luo2025bevplace++,
  title={Bevplace++: Fast, robust, and lightweight lidar global localization for unmanned ground vehicles},
  author={Luo, Lun and Cao, Si-Yuan and Li, Xiaorui and Xu, Jintao and Ai, Rui and Yu, Zhu and Chen, Xieyuanli},
  journal={IEEE Transactions on Robotics},
  year={2025},
  publisher={IEEE}
}

@article{grainge2024structured,
  title={Structured pruning for efficient visual place recognition},
  author={Grainge, Oliver and Milford, Michael and Bodala, Indu and Ramchurn, Sarvapali D and Ehsan, Shoaib},
  journal={IEEE Robotics and Automation Letters},
  year={2024},
  publisher={IEEE}
}

@inproceedings{he2017channel,
  title={Channel pruning for accelerating very deep neural networks},
  author={He, Yihui and Zhang, Xiangyu and Sun, Jian},
  booktitle={Proceedings of the IEEE international conference on computer vision},
  pages={1389--1397},
  year={2017}
}

@article{hinton2015distilling,
  title={Distilling the knowledge in a neural network},
  author={Hinton, Geoffrey and Vinyals, Oriol and Dean, Jeff},
  journal={arXiv preprint arXiv:1503.02531},
  year={2015}
}

@article{grainge2024design,
  title={Design space exploration of low-bit quantized neural networks for visual place recognition},
  author={Grainge, Oliver and Milford, Michael and Bodala, Indu and Ramchurn, Sarvapali D and Ehsan, Shoaib},
  journal={IEEE Robotics and Automation Letters},
  volume={9},
  number={6},
  pages={5070--5077},
  year={2024},
  publisher={IEEE}
}

@inproceedings{liu2023oscillation,
  title={Oscillation-free quantization for low-bit vision transformers},
  author={Liu, Shih-Yang and Liu, Zechun and Cheng, Kwang-Ting},
  booktitle={International conference on machine learning},
  pages={21813--21824},
  year={2023},
  organization={PMLR}
}

@inproceedings{jacob2018quantization,
  title={Quantization and training of neural networks for efficient integer-arithmetic-only inference},
  author={Jacob, Benoit and Kligys, Skirmantas and Chen, Bo and Zhu, Menglong and Tang, Matthew and Howard, Andrew and Adam, Hartwig and Kalenichenko, Dmitry},
  booktitle={Proceedings of the IEEE/CVF Conference on Computer Vision and Pattern Recognition},
  pages={2704--2713},
  year={2018}
}

@article{zhang2022quantifying,
  title={Quantifying the knowledge in a DNN to explain knowledge distillation for classification},
  author={Zhang, Quanshi and Cheng, Xu and Chen, Yilan and Rao, Zhefan},
  journal={IEEE Transactions on Pattern Analysis and Machine Intelligence},
  volume={45},
  number={4},
  pages={5099--5113},
  year={2022},
  publisher={IEEE}
}

@inproceedings{dosovitskiy2021image,
  title={An image is worth 16x16 words: Transformers for image recognition at scale},
  author={Dosovitskiy, Alexey},
  booktitle={International Conference on Learning Representations},
  year={2021}
}

@book{villani2008optimal,
  title={Optimal transport: old and new},
  author={Villani, C{\'e}dric and others},
  volume={338},
  year={2008},
  publisher={Springer}
}

@inproceedings{wang2023posediffusion,
  title={Posediffusion: Solving pose estimation via diffusion-aided bundle adjustment},
  author={Wang, Jianyuan and Rupprecht, Christian and Novotny, David},
  booktitle={Proceedings of the IEEE/CVF International Conference on Computer Vision},
  pages={9773--9783},
  year={2023}
}

@inproceedings{cheng2020explaining,
  title={Explaining knowledge distillation by quantifying the knowledge},
  author={Cheng, Xu and Rao, Zhefan and Chen, Yilan and Zhang, Quanshi},
  booktitle={Proceedings of the IEEE/CVF conference on computer vision and pattern recognition},
  pages={12925--12935},
  year={2020}
}

@article{yu2022lidar,
  title={LiDAR-based localization using universal encoding and memory-aware regression},
  author={Yu, Shangshu and Wang, Cheng and Wen, Chenglu and Cheng, Ming and Liu, Minghao and Zhang, Zhihong and Li, Xin},
  journal={Pattern Recognition},
  volume={128},
  pages={108685},
  year={2022},
  publisher={Elsevier}
}

@article{oquab2023dinov2,
  title={Dinov2: Learning robust visual features without supervision},
  author={Oquab, Maxime and Darcet, Timoth{\'e}e and Moutakanni, Th{\'e}o and Vo, Huy and Szafraniec, Marc and Khalidov, Vasil and Fernandez, Pierre and Haziza, Daniel and Massa, Francisco and El-Nouby, Alaaeldin and others},
  journal={arXiv preprint arXiv:2304.07193},
  year={2023}
}

@inproceedings{szegedy2015going,
  title={Going deeper with convolutions},
  author={Szegedy, Christian and Liu, Wei and Jia, Yangqing and Sermanet, Pierre and Reed, Scott and Anguelov, Dragomir and Erhan, Dumitru and Vanhoucke, Vincent and Rabinovich, Andrew},
  booktitle={Proceedings of the IEEE conference on computer vision and pattern recognition},
  pages={1--9},
  year={2015}
}

@inproceedings{belilovsky2020decoupled,
  title={Decoupled greedy learning of cnns},
  author={Belilovsky, Eugene and Eickenberg, Michael and Oyallon, Edouard},
  booktitle={International Conference on Machine Learning},
  pages={736--745},
  year={2020},
  organization={PMLR}
}

@article{duan2022training,
  title={Training deep architectures without end-to-end backpropagation: A survey on the provably optimal methods},
  author={Duan, Shiyu and Principe, Jose C},
  journal={IEEE Computational Intelligence Magazine},
  volume={17},
  number={4},
  pages={39--51},
  year={2022},
  publisher={IEEE}
}

@article{chen2023sc,
  title={SC$^2$-PCR++: Rethinking the Generation and Selection for Efficient and Robust Point Cloud Registration},
  author={Chen, Zhi and Sun, Kun and Yang, Fan and Guo, Lin and Tao, Wenbing},
  journal={IEEE transactions on pattern analysis and machine intelligence},
  volume={45},
  number={10},
  pages={12358--12376},
  year={2023},
  publisher={IEEE}
}

@inproceedings{wang2019deep,
  title={Deep closest point: Learning representations for point cloud registration},
  author={Wang, Yue and Solomon, Justin M},
  booktitle={Proceedings of the IEEE/CVF international conference on computer vision},
  pages={3523--3532},
  year={2019}
}

@inproceedings{komorowski2021minkloc3d,
  title={Minkloc3d: Point cloud based large-scale place recognition},
  author={Komorowski, Jacek},
  booktitle={Proceedings of the IEEE/CVF winter conference on applications of computer vision},
  pages={1790--1799},
  year={2021}
}

@inproceedings{uy2018pointnetvlad,
  title={Pointnetvlad: Deep point cloud based retrieval for large-scale place recognition},
  author={Uy, Mikaela Angelina and Lee, Gim Hee},
  booktitle={Proceedings of the IEEE conference on computer vision and pattern recognition},
  pages={4470--4479},
  year={2018}
}

@inproceedings{li2023sgloc,
  title={SGLoc: Scene geometry encoding for outdoor LiDAR localization},
  author={Li, Wen and Yu, Shangshu and Wang, Cheng and Hu, Guosheng and Shen, Siqi and Wen, Chenglu},
  booktitle={Proceedings of the IEEE/CVF Conference on Computer Vision and Pattern Recognition},
  pages={9286--9295},
  year={2023}
}

@inproceedings{yang2025raloc,
  title={RALoc: Enhancing Outdoor LiDAR Localization via Rotation Awareness},
  author={Yang, Yuyang and Li, Wen and Ao, Sheng and Xu, Qingshan and Yu, Shangshu and Guo, Yu and Zhou, Yin and Shen, Siqi and Wang, Cheng},
  booktitle={Proceedings of the IEEE/CVF International Conference on Computer Vision},
  pages={3304--3313},
  year={2025}
}

@inproceedings{yang2024lisa,
  title={Lisa: Lidar localization with semantic awareness},
  author={Yang, Bochun and Li, Zijun and Li, Wen and Cai, Zhipeng and Wen, Chenglu and Zang, Yu and Muller, Matthias and Wang, Cheng},
  booktitle={Proceedings of the IEEE/CVF Conference on Computer Vision and Pattern Recognition},
  pages={15271--15280},
  year={2024}
}

@inproceedings{chen2022dfnet,
  title={DFNet: Enhance Absolute Pose Regression with Direct Feature Matching},
  author={Chen, Shuai and Li, Xinghui and Wang, Zirui and Prisacariu, Victor},
  booktitle={Proceedings of the European Conference on Computer Vision},
  year={2022}
}

@inproceedings{kendall2017geometric,
  title={Geometric loss functions for camera pose regression with deep learning},
  author={Kendall, Alex and Cipolla, Roberto},
  booktitle={Proceedings of the IEEE conference on computer vision and pattern recognition},
  pages={5974--5983},
  year={2017}
}

@inproceedings{kendall2015posenet,
  title={Posenet: A convolutional network for real-time 6-dof camera relocalization},
  author={Kendall, Alex and Grimes, Matthew and Cipolla, Roberto},
  booktitle={Proceedings of the IEEE international conference on computer vision},
  pages={2938--2946},
  year={2015}
}

@inproceedings{moreau2022lens,
  title={Lens: Localization enhanced by nerf synthesis},
  author={Moreau, Arthur and Piasco, Nathan and Tsishkou, Dzmitry and Stanciulescu, Bogdan and de La Fortelle, Arnaud},
  booktitle={Conference on Robot Learning},
  pages={1347--1356},
  year={2022},
  organization={PMLR}
}

@article{shavit2023coarse,
  title={Coarse-to-fine multi-scene pose regression with transformers},
  author={Shavit, Yoli and Ferens, Ron and Keller, Yosi},
  journal={IEEE transactions on pattern analysis and machine intelligence},
  volume={45},
  number={12},
  pages={14222--14233},
  year={2023},
  publisher={IEEE}
}

@inproceedings{wang2020atloc,
  title={Atloc: Attention guided camera localization},
  author={Wang, Bing and Chen, Changhao and Lu, Chris Xiaoxuan and Zhao, Peijun and Trigoni, Niki and Markham, Andrew},
  booktitle={Proceedings of the AAAI Conference on Artificial Intelligence},
  pages={10393--10401},
  year={2020}
}

@article{chen2024binarized,
  title={Binarized diffusion model for image super-resolution},
  author={Chen, Zheng and Qin, Haotong and Guo, Yong and Su, Xiongfei and Yuan, Xin and Kong, Linghe and Zhang, Yulun},
  journal={Advances in Neural Information Processing Systems},
  volume={37},
  pages={30651--30669},
  year={2024}
}

@inproceedings{xu2022ida,
  title={Ida-det: An information discrepancy-aware distillation for 1-bit detectors},
  author={Xu, Sheng and Li, Yanjing and Zeng, Bohan and Ma, Teli and Zhang, Baochang and Cao, Xianbin and Gao, Peng and L{\"u}, Jinhu},
  booktitle={European Conference on Computer Vision},
  pages={346--361},
  year={2022},
  organization={Springer}
}

@INPROCEEDINGS{pathfinder,
  author={Yin, Kaijie and Gao, Tian and Kong, Hui},
  booktitle={IEEE/RSJ International Conference on Intelligent Robots and Systems}, 
  title={Pathfinder for Low-altitude Aircraft with Binary Neural Network}, 
  year={2025},
  volume={},
  number={},
  pages={20692-20699}
}

@inproceedings{barnes2020oxford,
  title={The oxford radar robotcar dataset: A radar extension to the oxford robotcar dataset},
  author={Barnes, Dan and Gadd, Matthew and Murcutt, Paul and Newman, Paul and Posner, Ingmar},
  booktitle={IEEE international conference on robotics and automation},
  pages={6433--6438},
  year={2020},
  organization={IEEE}
}

@inproceedings{shen2021s2,
  title={S2-bnn: Bridging the gap between self-supervised real and 1-bit neural networks via guided distribution calibration},
  author={Shen, Zhiqiang and Liu, Zechun and Qin, Jie and Huang, Lei and Cheng, Kwang-Ting and Savvides, Marios},
  booktitle={Proceedings of the IEEE/CVF conference on computer vision and pattern recognition},
  pages={2165--2174},
  year={2021}
}

@inproceedings{chen2021wasserstein,
  title={Wasserstein contrastive representation distillation},
  author={Chen, Liqun and Wang, Dong and Gan, Zhe and Liu, Jingjing and Henao, Ricardo and Carin, Lawrence},
  booktitle={Proceedings of the IEEE/CVF conference on computer vision and pattern recognition},
  pages={16296--16305},
  year={2021}
}

@article{cuturi2013sinkhorn,
  title={Sinkhorn distances: Lightspeed computation of optimal transport},
  author={Cuturi, Marco},
  journal={Advances in neural information processing systems},
  volume={26},
  year={2013}
}

@article{carlevaris2016university,
  title={University of Michigan North Campus long-term vision and lidar dataset},
  author={Carlevaris-Bianco, Nicholas and Ushani, Arash K and Eustice, Ryan M},
  journal={The International Journal of Robotics Research},
  volume={35},
  number={9},
  pages={1023--1035},
  year={2016},
  publisher={Sage Publications Sage UK: London, England}
}

@inproceedings{cho2019efficacy,
  title={On the efficacy of knowledge distillation},
  author={Cho, Jang Hyun and Hariharan, Bharath},
  booktitle={Proceedings of the IEEE/CVF international conference on computer vision},
  pages={4794--4802},
  year={2019}
}

@article{victor2023safety,
  title={Safety performance of the Waymo rider-only automated driving system at one million miles},
  author={Victor, Trent and Kusano, Kristofer and Gode, Tilia and Chen, Ruoshu and Schwall, Matthew},
  journal={Tech. Rep.},
  year={2023}
}

@inproceedings{sun2020scalability,
  title={Scalability in perception for autonomous driving: Waymo open dataset},
  author={Sun, Pei and Kretzschmar, Henrik and Dotiwalla, Xerxes and Chouard, Aurelien and Patnaik, Vijaysai and Tsui, Paul and Guo, James and Zhou, Yin and Chai, Yuning and Caine, Benjamin and others},
  booktitle={Proceedings of the IEEE/CVF conference on computer vision and pattern recognition},
  pages={2446--2454},
  year={2020}
}

@inproceedings{wolcott2015fast,
  title={Fast LIDAR localization using multiresolution Gaussian mixture maps},
  author={Wolcott, Ryan W and Eustice, Ryan M},
  booktitle={2015 IEEE international conference on robotics and automation (ICRA)},
  pages={2814--2821},
  year={2015},
  organization={IEEE}
}

@InProceedings{Le_2023_CVPR,
    author    = {Le, Phuoc-Hoan Charles and Li, Xinlin},
    title     = {BinaryViT: Pushing Binary Vision Transformers Towards Convolutional Models},
    booktitle = {Proceedings of the IEEE/CVF Conference on Computer Vision and Pattern Recognition (CVPR) Workshops},
    month     = {June},
    year      = {2023},
    pages     = {4665-4674}
}

@inproceedings{touvron2021training,
  title={Training data-efficient image transformers \& distillation through attention},
  author={Touvron, Hugo and Cord, Matthieu and Douze, Matthijs and Massa, Francisco and Sablayrolles, Alexandre and J{\'e}gou, Herv{\'e}},
  booktitle={International conference on machine learning},
  pages={10347--10357},
  year={2021},
  organization={PMLR}
}

@article{wang2021pointloc,
  title={Pointloc: Deep pose regressor for lidar point cloud localization},
  author={Wang, Wei and Wang, Bing and Zhao, Peijun and Chen, Changhao and Clark, Ronald and Yang, Bo and Markham, Andrew and Trigoni, Niki},
  journal={IEEE Sensors Journal},
  volume={22},
  number={1},
  pages={959--968},
  year={2021},
  publisher={IEEE}
}

@article{yu2022stcloc,
  title={Stcloc: Deep lidar localization with spatio-temporal constraints},
  author={Yu, Shangshu and Wang, Cheng and Lin, Yitai and Wen, Chenglu and Cheng, Ming and Hu, Guosheng},
  journal={IEEE Transactions on Intelligent Transportation Systems},
  volume={24},
  number={1},
  pages={489--500},
  year={2022},
  publisher={IEEE}
}

@inproceedings{yin2024si,
  title={Si-bivit: Binarizing vision transformers with spatial interaction},
  author={Yin, Peng and Zhu, Xiaosu and Song, Jingkuan and Gao, Lianli and Shen, Heng Tao},
  booktitle={Proceedings of the 32nd ACM International Conference on Multimedia},
  pages={8169--8178},
  year={2024}
}

@article{chetlur2014cudnn,
  title={cudnn: Efficient primitives for deep learning},
  author={Chetlur, Sharan and Woolley, Cliff and Vandermersch, Philippe and Cohen, Jonathan and Tran, John and Catanzaro, Bryan and Shelhamer, Evan},
  journal={arXiv preprint arXiv:1410.0759},
  year={2014}
}

@article{li2020accelerating,
  title={Accelerating binarized neural networks via bit-tensor-cores in turing gpus},
  author={Li, Ang and Su, Simon},
  journal={IEEE Transactions on Parallel and Distributed Systems},
  volume={32},
  number={7},
  pages={1878--1891},
  year={2020},
  publisher={IEEE}
}

\end{document}